\RequirePackage{fix-cm}

\documentclass[smallextended]{svjour3}       
\smartqed  
\usepackage{graphicx}
\usepackage{fancyhdr}
\usepackage{etoolbox}
\usepackage[export]{adjustbox}

\usepackage{float}
\usepackage{amsmath}

\usepackage{booktabs,chemformula}
\usepackage{hyperref}
\usepackage[numbers]{natbib}

\usepackage{scrextend}
\usepackage{etoolbox}

\usepackage{etoolbox}
\newcommand*{\affaddr}[1]{#1} 
\newcommand*{\affmark}[1][*]{\textsuperscript{#1}}

\usepackage{lineno}

\begin{document}

\title{Capturing Stance Dynamics in Social Media:\\Open Challenges and Research Directions}

\titlerunning{Dynamics in Stance Detection }        

\author{%
Rabab Alkhalifa \protect\affmark[1,2] \and  Arkaitz Zubiaga \affmark[1]}
\authorrunning{Alkhalifa \and Zubiaga}

\institute{
Rabab Alkhalifa \at 
 \email{raalkhalifa@iau.edu.sa}     \\      
 \emph{Present address:} \email{r.a.a.alkhalifa@qmul.ac.uk}
           \and
            Arkaitz Zubiaga \at
 \email{a.zubiaga@qmul.ac.uk} \\
 \emph{Website:} \url{http://www.zubiaga.org/} \\
\affaddr{\affmark[1]School of Electronic Engineering and Computer Science, Queen Mary University of London, Mile End Road, London, E1 4NS, UK}\\
\affaddr{\affmark[2]College of Computer Science and Information Technology, Imam Abdulrahman Bin Faisal University, P.O.Box 1982, Dammam 31451, Kingdom of Saudi Arabia}\\
}

\maketitle


\begin{abstract}
Social media platforms provide a goldmine for mining public opinion on issues of wide societal interest and impact. Opinion mining is a problem that can be operationalised by capturing and aggregating the stance of individual social media posts as supporting, opposing or being neutral towards the issue at hand. While most prior work in stance detection has investigated datasets that cover short periods of time, interest in investigating longitudinal datasets has recently increased. Evolving dynamics in linguistic and behavioural patterns observed in new data require adapting stance detection systems to deal with the changes. In this survey paper, we investigate the intersection between computational linguistics and the temporal evolution of human communication in digital media. We perform a critical review of emerging research considering dynamics, exploring different semantic and pragmatic factors that impact linguistic data in general, and stance in particular. We further discuss current directions in capturing stance dynamics in social media. We discuss the challenges encountered when dealing with stance dynamics, identify open challenges and discuss future directions in three key dimensions: utterance, context and influence.

\end{abstract}

\section{Introduction}
\label{intro}

With the proliferation of social media and blogs that enable anyone to post and share content, professional accounts from news organisations and governments aren't any longer the sole reporters of events of public interest \cite{kapoor2018advances}. Posting a tweet or a video, or writing an article that goes viral and reaches millions of individuals is now more accessible to ordinary citizens \cite{mills2012virality}. Where anyone can post their views on social media, the use of social media gains ground as a data source for public opinion mining. This data source provides a goldmine for nowcasting public opinion by aggregating the stances expressed by individual social media posts on a particular issue.

Research in stance detection has recently attracted an increasing interest \cite{Kucuk2019}, with two main directions. One of the directions includes determining the stance of posts as supporting, denying, querying or commenting on a rumour, which is used as a proxy to predict the likely veracity of the rumour in question \cite{Zubiaga2016,zubiaga2018discourse,hardalov2021survey}. The other direction, which is the focus of this paper, defines stance detection as a three-way classification task where the stance of each post is one of supporting, opposing or neutral \cite{augenstein2016stance}, indicating the viewpoint of a post towards a particular issue. This enables mining public opinion as the aggregate of stances of a large collection of posts.

In using stance detection to mine public opinion, most research has been operationalised by evaluating on temporally constrained datasets. This presents important limitations when one wants to apply the models on temporally distant test datasets, as recent studies demonstrate. Due to the rapidly evolving nature of social media content, as well as the rapid evolution of people's opinions, a model trained on an old dataset may not perform at the same level on new data \cite{alkhalifa2021opinions}. This review paper discusses the different factors that impact evolving changes in public opinion and their impact on stance detection models, discussing previous work studying this problem.

This paper reviews research on stance detection from an interdisciplinary perspective focusing on the impact of time on model performance. We present current progress in addressing these factors, discuss existing datasets with their potential and limitations for investigating stance dynamics, as well as identify open challenges and future research directions. We focus particularly on the dynamic factors impacting stance, including variations across cultures and regions, but also temporal changes caused by events in the real world leading to changes in public opinion. We set forth directions for future research with the aim of furthering the consideration of dynamics surrounding the stance detection task. Our goal is to draw a framework which links the current trends in stance detection, from an interdisciplinary perspective covering computational challenges bridging with broader linguistics and social science angles.



\subsection{Overview}

Linguists are interested in understanding human language, which is often dependent on its context \cite{Englebretson2007}. The ethnographic definition of stance in everyday language may vary from the academic definition of stance given in the literature \cite{Englebretson2007}. Consequently, the definition of stance can be analysed from different perspectives, while most NLP work tends to focus on one of them. The prevalent definition of stance in NLP research stems from a usage-based perspective defined in the field of linguistics and is described by \citet{Englebretson2007} in which stance is dependent on personal belief, evaluation or attitude. Additionally, stance can be seen as the expression of a viewpoint and it relates to the analysis and interpretation of written or spoken language using lexical, grammatical and phonetic characteristics \cite{Cossette}. For example, everyday words or phrases used by people during working hours or in performing specific tasks can express subjective features \cite{Cossette} which can be used by NLP researchers in different applications. However, the stance term may appear and be used differently by researchers as it is strongly relevant to one's own interpretation of the concept.

\subsection{Computational View of Stance Detection}



Stance, as a message conveying the point of view of the communicator, is the opinion from whom one thing is discovered or believed. As a computational task, stance detection is generally defined as that in which a classifier needs to determine if an input text expresses a supporting, neutral or opposing view \cite{Aldayel2019}. It is framed as a supervised classification task, where labelled instances are used to train a classification model, which is then applied on unseen test data.

While humans can easily infer whether an author is in favour or against a specific event, the task becomes more challenging when performed at scale, due to the need for automated NLP methods. Consequently, the stance detection task has attracted an increasing interest in the scientific community, including scholars from linguistics and communication as well as computational linguistics. However, the need to automate the task by means of NLP methods is still in its infancy with a growing body of ongoing research.

Understanding stance expressed in text is a critical, yet challenging task and it is the main focus of this review paper. Stance is often implicit and needs to be inferred rather than directly determined from explicit expressions given in the text; indeed the target may not be directly mentioned \cite{Somasundaran2009,Mohammad2016b}. However, given the scale of social media data, understanding attitudes and responses of people to different events becomes unmanageable if done manually. Current stance detection approaches leverage machine learning and NLP models to study political and other opinionated issues \cite{Volkova2016,Al-Ayyoub2018,Dandrea2019,Johnson2016,Lai2017}. However, using persuasive writing techniques and word choices \cite{BURGOON1975} to convey a stance brings important challenges for current state-of-the-art models as there is a need to capture these features in a large-scale dataset. Recent research is increasingly considering pragmatic factors in texts, adopting stance dynamics and the impact of language evolution. Research in this direction can shed light into other dimensions when defining and analysing stance. However, building representations for complex, shifting or problematic meanings is still an open problem that needs exploration.

\subsection{Capturing Dynamics in Stance Detection}
\label{sec:taxonomy}

The stance detection task overlaps with, and is closely related to, different classification tasks such as sentiment analysis \cite{Chakraborty2020}, troll detection \cite{Tomaiuolo2020}, rumour and fake news detection \cite{zubiaga2018detection,rani2020rumour,collins2020fake}, and argument mining \cite{Lawrence2020}. In addition, stance can be impacted by the discursive and dynamic nature of the task \cite{Mohammad2017a,Somasundaran2009,Simaki2017a}.

\begin{figure}
 \begin{center}
  \includegraphics[width=1\textwidth]{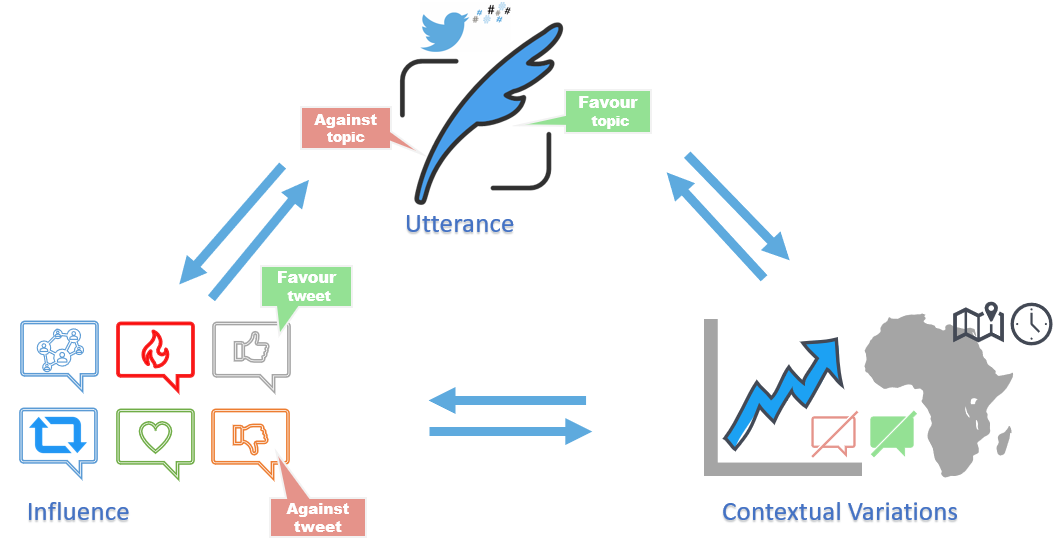}
  \caption{Three dimensions around stance detection.}
  \label{fig:dimensions}
 \end{center}
\end{figure}

In reviewing the literature on stance dynamics, we break down our review into three different dimensions (see Figure \ref{fig:dimensions}), which cover the different aspects impacting how stance is formed and how it evolves:
\begin{itemize}
 \item \textbf{stance utterance}, referring to a single message conveying a particular stance towards a target.
 \item \textbf{stance context}, referring to the pragmatic, spatiotemporal and diachronic factors that make stance an evolving phenomenon.
 \item \textbf{stance influence}, referring to social factors including the author of a post, as well as reactions towards, and activity around, messages expressing a particular stance.
\end{itemize}

In what follows, we delve into each of these dimensions and associated literature.

\section{Stance Utterance}
\label{ssec:stance-utterance}

Stance utterance refers to the stance expressed in a single message \cite{mohammad-etal-2016-semeval}, and reflects human interpretation of an event. It represents the features that form the textual viewpoint and they are essential for human inference and interpretation. Textual data can be analysed based on different features, which previous work have tackled by looking at a range of different challenges, which we discuss next.

One of the challenges in detecting the stance of a single utterance is \textbf{target identification}, i.e. determining who or what the stance is referring to. For example, in the utterance ``I am supportive of A, but I'm totally against B'', the author expresses a supporting stance towards target A and an opposing stance towards target B. The \textbf{target} may be implicit and not always directly mentioned in the text \cite{Schaefer2019a}. The target may be implicitly referred to \cite{Sobhani2019a}, or only aspects of it may be mentioned \cite{Bar-Haim}. These cases present the additional challenge of having to detect the target being referred to in a text prior to detecting the stance. Retrieval of messages likely referring to a target, as a first step to then do the target identification, is a challenge. Achieving high recall in relevant message retrieval can be difficult in the case of implicit messages \cite{Mohammad2017a}. In addition, there is a risk of detecting false positives where a message may not be about the target at all, may not contain data expressing a stance, or may hold multiple stances in the same utterance \cite{Lai2019c,Simaki2017a}.

Nuances in the \textbf{wording} of an utterance can present another challenge in detecting stance. Opinions are not always explicitly expressed, and can also be implicit, explicit, ironic, metaphoric, uncertain, etc. \cite{Sun2016,Al-Ayyoub2018,Simaki2018c}, which make stance detection more challenging. Moreover, surrounding words and symbols can alter the stance of an utterance \cite{Sun2016}, e.g. negating words or ironic emojis inverting the meaning of a text, which are especially challenging to detect. Recent models increasingly make use of more sophisticated linguistic and contextual features to infer stance from text. For example, looking at the \textbf{degree of involvement} by using special lexical terms, e.g. slang, jargon, specialist terms, and the informal lexicons associated with social intimacy \cite{Tausczik2010,Hamilton2016a,Rumshisky2017,Liu2015}. Also, use of \textbf{embeddings} where concept meanings can be biased and highly impacted by the cultural background and beliefs may lead to varying interpretations \cite{Shoemark,Kutuzov2018,Hamilton2016,Dubossarsky2017,Xu2019c}. 

The \textbf{framing} of an utterance can also play an important role in the detection of stance. Framing refers to the adaptation of the wording to convey a specific interpretation of a story to a targeted audience \cite{Walker2012} as in language and word choices, this can be seen in the following forms:
 
\begin{itemize}
 \item \textbf{Reasoning/supporting evidence} about the target or aspects of it \cite{Hasan2014,Addawood2016,Bar-Haim,Simaki2017a}. For example, \citet{Bar-Haim} define the claim stance classification task as consisting of a target, a set of claims and the stance of these claims as either supporting or opposing the target. Further, they simplify the task by looking for sentiment and contrast meaning between a given pair of target phrase and the topic candidate anchor phrase.

 \item \textbf{Attitude}: using single lexical terms holding polarity features related to the author sentiment and reaction to an event. Such word choices can be positive, negative, offensive, harmful, suspicious, aggressive, extremist \cite{blvstak2017machine}. For instance, emotion and sentiment expressed in the text \cite{Deitrick2013,Xu2011}, which may express the author's view on the importance (or lack thereof) of the target.

 \item \textbf{Persuasion and quality of communication} \cite{Hamilton2015,Aune1993} through using grammatically correct sentences. For example, \citet{Lai2019c} concluded that people connected to users taking cross-stance attitudes become less polarised and use neutral style when expressing their stance.
\end{itemize}

Another challenge in stance detection is determining if \textbf{stance is present} in an utterance, as the text may be neutral and not be opinionated towards the target. This involves determining whether or not an utterance expresses a stance, and subsequently determining the type of stance the author is taking \cite{Mohammad2016b,Zubiaga2016,Simaki2017a}. For example, the most basic way of stance-taking could be the more extreme positions such as in favour or against to less extreme positions such as asserting, questioning, responding, commanding, advising, and offering which may lead to conversational and threaded stance context \cite{Zubiaga2016}. In such a context stance moves from its singular utterance structure to its augmenting component (see Section \ref{ssec:stance-influence}).

\section{Stance Context}
\label{ssec:contextual-variations}

Stance context refers to the impact of external factors in an utterance, including the collective viewpoint of a society in relation to the interpretation of a particular target. Context aims to capture the stance occurring in longitudinally evolving contexts, and can be impacted by shifts in opinions over time, locations, or cultures, among others. Any stance inference requires consideration of other points of view, potential stereotypes, as well as how public opinion evolved over time. This represents our understanding of the world dynamics and how stance may change over time \cite{Lai2019c,Volkova2016}. Stance towards a topic may be considered stable only when it has the same polarity over time. Context involves factors causing changes in public opinion over time, such as real world events. Context constructs complex, shifting or problematic meanings which change the entire view of an event \cite{Azarbonyad2017,Stewart2017}. We discuss the two main aspects that are considered when modelling stance context, which include \textbf{spatiotemporal changes} and \textbf{social changes}.

Collective and individual stance towards a target can be impacted by \textbf{spatiotemporal factors} \cite{Volkova2016,Jackson2019}. Events occurring in different locations/times get different attention depending on how likely they are to happen again and how unusual they are \cite{baly2018predicting,hamborg2019automated}. Consequently, the audience judging the events have their own biases depending on the cultural and ideological background, which leads to variations in stance across regions.
 
Even when we restrict geographical locations, there are other factors leading to \textbf{social changes} that have an impact on public opinion and stance. Social changes around a topic can lead to shifts in opinions \cite{Volkova2016,Dandrea2019,Lai2018,Lai2019c}. This can pose a significant challenge, particularly with the tendency in NLP to using distributed representations of words driven by co-occurrence frequency of words using sliding windows, and considering polysemy in more advanced language models. Words are treated based on their contextual similarity rather than solely based on their isolated frequencies. In order to build these models, one needs large collections of documents with a diverse vocabulary to produce high quality vector representations for different words. Consequently, these models rely on the amount of training data available, and the dimensionality of the word vectors \cite{Mikolov2013a}. The emergence of out of vocabulary (OOV) words, not seen by these models, can be one of its main limitations. Different methods have been proposed to mitigate these limitations, for example through character-level representations in ELMo or FastText, and sub-word representations in BERT allowing models to incorporate segmented representations for unseen words \cite{ha2020improving}. In prediction models, character-level and subword representations can lead to performance improvements with a trade-off on reduced model explainability; ongoing research is however investigating how to improve model explainability, exploiting for example attention scores produced by BERT \cite{bodria2020explainability}. Moreover, newly emerging words or words that shift their meaning over time would lead to outdated models. Challenges relating to social changes can be further broken down into the following:

\begin{itemize}
 \item \textbf{Linguistic shift}, which is defined as slow and regular changes in the core meaning of a word. For example, ``the word GAY shifting from meaning CAREFREE to HOMOSEXUAL during the 20th century" \cite{Kutuzov2018}. This is also reflected in the semantic meaning of emoticons across different contexts, languages and cultures \cite{walid2018Emoji}. In multilingual settings, code-mixing of two languages in the same utterance \cite{khanuja2020gluecos}, or borrowing a word from a different language due to influence from other languages, rather than internal changes in the same language.
   
 \item \textbf{Usage change}, which is the local change of a word’s nearest semantic neighbours from one meaning to another, as in the shift of word the ``prison CELL to CELL phone" which is more of a cultural change than a semantic change \cite{Hamilton2016}. Thus, different viewpoints allow collective stance to change especially when a story is viewed through different eyes and interpreted differently. For example, focusing on UK politics, \citet{Azarbonyad2017} revealed that ``The meaning given by Labours to MORAL is shifted from a PHILOSOPHICAL concept to a LIBERAL concept over time. In the same time, the meaning of this word is shifted from a SPIRITUAL concept to a RELIGIOUS concept from the Conservatives’ viewpoint. Moreover, two parties gave very different meanings to this word. Also, the meaning of DEMOCRACY is stable over time for both parties. However, Conservatives refer to democracy mostly as a UNITY concept, while Labours associate it with FREEDOM and SOCIAL JUSTICE."
  
 \item \textbf{Changes in cultural associations}, which is measured as the distance between two words in the semantic space, as in ``IRAQ or SYRIA being associated with the concept of WAR after armed conflicts had started in these countries" \cite{Kutuzov2018}. Also, the type of sentiment bore by a  word can change over time. For example, ``the word SLEEP acquiring more negative connotations related to sleep disorders, when comparing its 1960s contexts to its 1990s contexts'' \cite{Gulordava2011}. Moreover, studies have also looked at the relatedness of words over time, by looking at how the strength of the association between words changes. For example, \citet{Rosin2017} introduced a relationship model that supports the task of identifying, given two words (e.g. Obama and president), when they relate most to each other, having longitudinal data collections as input.
  
\end{itemize}
  
\section{Stance Influence}
\label{ssec:stance-influence}

Stance influence refers to the aggregated importance surrounding an individual message expressing a particular stance, and can be measured by using different qualitative and quantitative metrics. These include the author's profile and others' reactions to a message.

Influence defines the quality of an utterance to make an impact, and can vary depending on the popularity and reputation of the author, as well as the virality of a post, among others. Next we discuss three aspects which are relevant to stance influence, i.e. \textbf{threading comments}, \textbf{network homophily} and \textbf{author profile}.

Social media platforms provide a place for conversations to develop, which lead to \textbf{threaded conversations} or \textbf{tree-structured conversations}. The formation of these conversations enables exchanging viewpoints on top of the initial author's stance \cite{Zubiaga2016,Guerra2017,Lai2019c}. For example, \citet{Lai2019c} observed that users make use of replies for expressing diverging opinions. Research looking at whether retweeting a post indicates endorsement is so far inconsistent. \citet{Lai2018} observe that people tend to retweet what they agree on. Conversely, \cite{Guerra2017} argued that a retweet does not indicate supporting its underlying opinion.

There is evidence showing that social media users tend to connect and interact with other like-minded users \cite{Lai2017a,Conover2012}, which is also known as the phenomenon of \textbf{network homophily}. \citet{Lai2019c} looked at the impact of different characteristics of social media in sharing stance, showing for example that opposing opinions generally occur through replies as rather than through retweets or quotes, polarisation varies over time, e.g. increasing in the proximity of elections.

The identity of the person posting a piece of text expressing a stance, or the \textbf{author's profile}, can also play an important role in the development of stance, for example if an influential user expresses an opinion. Two key factors of an author's profile include:

\begin{itemize}
 \item \textbf{Author's ideology and background}, often inferred by observing the user's profile \cite{elfardy2016cu,Conover2012,Yan2018a,Lai2019c}, can be used as additional features to determine the stance expressed by a user, rather than solely using the textual content of a post \citet{mohammad-etal-2016-semeval,Mohammad2016b}.
 \item \textbf{Author's stance in the temporal space} \cite{Garcia2015}. For instance, media organisations may express viewpoints through different frames \cite{Hasan2014}, which takes time to be assessed \cite{Zubiaga2016} and may also impact stance evolution and people's stand points. This can also have an impact on how threaded conversations are developed.
\end{itemize}

\section{Datasets}
\label{sec:datasets}

To study the stance detection task, different datasets covering various topics and pragmatic aspects have been created by researchers. 

Table \ref{tab:datasets} shows the list of stance datasets available, along with their key characteristics; these include the time frame they cover, a key aspect in our focus on stance dynamics, as we are interested in identifying the extent to which existing datasets enable this analysis. For ten of the datasets we found, the time frame covered by the data is not indicated (marked in the table as N/A), which suggests that temporal coverage was not the main focus of these works. The rest of the datasets generally cover from a month to a maximum of 1 or 2 years; while the latter provides some more longitudinal coverage, we argue that it is not enough to capture major societal changes. The exception providing a dataset that covers a longer period of time is that by \citet{Conforti2020WTWT} and \citet{Addawood2017}, with five years' worth of data. 

Despite the availability of multiple stance datasets and their ability to solve different generalisability problems (e.g. across targets, languages and domains), this analysis highlights the need for more longitudinal datasets that would enable persistence for temporal stance detection and temporal adaptation, ideally across cultures and languages. For the few datasets that contain some degree of longitudinal content, such as \cite{Conforti2020WTWT} covering 57 months and \cite{Addawood2017} covering 61 months, the available data is sparsely distributed throughout the entire time period. This again urges the need for more longitudinal datasets, which in turn provide more density for each time period. While data labelling is expensive and hard to afford at scale, possible solutions may include use of distant supervision \cite{purver2012experimenting} for data collection and labelling or labelling denser datasets for specific time periods which are temporally distant from each other, despite leaving gaps between the time periods under consideration. Distant supervision has been widely used for other tasks such as sentiment analysis \cite{go2009twitter}, leading to datasets covering in some cases over 7 years \cite{yin2021emojification}, however its applicability to stance detection has not been studied as much.

In summary, we observe that existing datasets provide limited resources to capture language dynamics and leverage longitudinal analysis, which would then give rise to more research aiming to capture stance dynamics.

\begin{table}[htb]
\centering
\begin{adjustbox}{max width=\textwidth}
\begin{tabular}{ p{2.5cm} |l |c |l |l |l }
\toprule
\textbf{ref.} & \textbf{time frame}  & \textbf{months} & \textbf{topics (\#)}   & \textbf{source} & \textbf{language}  \\
\hline
\hline
\citet{Conforti2020WTWT} & Apr 2014 - Dec 2018 & 57 & Finance (2) & Twitter & English \\
\hline
\citet{Addawood2017} & Jan 2012 - & 61 & Women to Drive & Twitter & Arabic \\
                  & Jan 2017 &  & Movement (1) &    &    \\
\hline
\citet{Rajadesingan2014Opposing} & Apr 2013 & 1 & US Issues (1) & Twitter & English \\
\hline
\citet{Volkova2016} & Sep 2014 - & 7 & Politics (1) & VKontakte & Russian \\
                  & Mar 2015 &    &    &    & Ukrainian \\
\hline
\citet{Zubiaga2016} & Aug 2014 - & 15 & News events (9) & Twitter & English \\
                  & Oct 2015 &    &    &    & German \\
\hline
\citet{Mohammad2016b} & Jul 2015 & 1 & US Issues (6) & Twitter & English \\
\hline
\citet{Schuff2017} & Jul 2015 & 1 & US Issues (6) & Twitter & English \\
\hline
\citet{Simaki2017a}     & Jun - Aug 2015 & 3 & Political blogs(1) & the BBC &  English \\
\hline
\citet{Kucuk2019} & Aug - Sep 2015 & 2 & Sport (2) & Twitter & Turkish \\
\hline
\citet{Sobhani2019a} & Oct 2015 - & 5 & US Issues (4) & Twitter & English \\
                  & Feb 2016 &    &    &    &    \\
\hline
\citet{Addawood2016}  & Jan - Mar 2016 & 3 & Products (1) & Web & English \\
\hline
\citet{Lai2019c,Lai2018} & Nov - Dec 2016 & 2 & Italian Referend. (1) & Twitter & Italian \\
\hline
\citet{Yan2018a} & Jan - Dec 2016  & 12 & US Issues (2) & Twitter & English \\
\hline
\citet{Dandrea2019} & Sep 2016 - & 10 & Health (1) & Twitter & Italian \\
                  & Jun 2017 &    &    &    &    \\
\hline
\citet{Lozhnikov2020} & Nov 2017     & 1 & Politics(1) & Twitter & Russian \\
                  &    &    &    & Meduza &    \\
                  &    &    &    & Russia Today &     \\
\hline
\citet{Baly2018b} & Jan 2016 - & 12 & Middle East(1) & News articles & Arabic \\
                  & Dec 2017 &    &    &    &    \\
\hline
\citet{Somasundaran2009} & N/A     & N/A & Products (1) & Convinceme & English \\
\hline
\citet{Anand2011} & N/A     & N/A & Politics (12) & Convinceme & English \\
\hline
\citet{Walker2012a} & N/A     & N/A & US Issues (12)     & 4forums & English \\
                  &    &    &     & Createdebate &    \\
\hline
\citet{skanda2017detecting} & N/A     & N/A & Indian Issues (4) & Facebook & Indian  \\
\hline
\citet{cicling2018} & N/A     & N/A & Czech Issues (2) & News & Czech \\ 
\hline
\citet{Xu2016a} & N/A     & N/A & Different topics (7) & Sina Weibo & Chinese \\
\hline
\citet{Ferreira2016} & N/A     & N/A & News Articles (1) & News articles & English \\
\hline
\citet{Hasan2014} & N/A     & N/A & US Issues (4) & Createdebate & English \\
\hline
\citet{Bar-Haim} & N/A     & N/A & Open domain  & IBM dataset & English \\
\hline
\citet{Taule2018} & N/A    & N/A & Catalan & Twitter & Spanish \\
                  &    &    & Independence (1) &    & Catalan \\

\bottomrule
\end{tabular}
\end{adjustbox}
\caption{Stance detection datasets, including the time frame covered.}
\label{tab:datasets}
\end{table}

\section{Open Challenges and Future Directions}
\label{sec:open-challenges}

In the previous sections we have discussed the three key factors relevant to stance and impacting its formation and temporal evolution, as well as existing datasets. In what follows, we discuss the main research challenges and set forth a number of future research directions. We first discuss core challenges, which are specific to stance detection, followed by general challenges, which are broader challenges that also have an impact on stance detection.

\subsection{Core Challenges}
\label{ssec:core-challenges}

There are numerous open challenges that are specific to the stance detection task. To the best of our knowledge, few studies have specifically focused on the \textbf{evolving nature of topics} and its impact on stance detection models. Moreover, fluctuation of word frequencies and distributions over time highlight both the challenge and the importance of the task. Commonalities between the source and target tasks tend to be crucial for successful transfer \cite{Vu2020NLP_Transfer}. However, recent NLP models have shifted to transfer learning and domain adaptation where target tasks contain limited training data \cite{xu2019adversarial}, source data pertains to a different domain \cite{zhang2020} or to a different language \cite{lai2020multilingual}. We anticipate two main directions that would help extend this research: (1) furthering research in transfer learning that looks more into transferring knowledge over time, as opposed to the more widely studied subareas looking into domain adaptation \cite{ramponi2020neural} or cross-lingual learning \cite{lin2019choosing}, and (2) increasing the availability of longitudinal datasets that would enable further exploration of temporal transfer learning.

The majority of existing datasets are from the \textbf{domain} of politics and to a lesser extent business, and are hence constrained in terms of topics. Broadening the topics covered in stance datasets should be one of the key directions in future research.

In general, existing datasets cover \textbf{short time spans} in languages including English \cite{Ferreira2016,Mohammad2016b,Simaki2017a,Somasundaran2009,Somasundaran2009,cicling2018,Walker2012a,Anand2011,Conforti2020WTWT}, Arabic \cite{Baly2018b,Addawood2017}, Italian \cite{Lai2018}, Chinese \cite{Xu2016a}, Turkish \cite{Kucuk2019}, Spanish and Catalan \cite{Taule2018}, Kannada \cite{skanda2017detecting}, German \cite{Zubiaga2016}, Russian \cite{Lozhnikov2020}.
Recent efforts in multilingual stance classification have also published datasets including German, French and Italian \cite{Mohtarami2019,Vamvas2020}, and English, French, Italian, Spanish and Catalan \cite{lai2020multilingual}, but are still limited in terms of the time frame covered. Longitudinal datasets annotated for stance would enable furthering research in this direction by looking into the temporal dynamics of stance.

The \textbf{quality and persistence of the data} are also important challenges that need attention. Annotation of stance is particularly challenging where a single post may contain multiple targets, or where users change their own stances towards a particular target, i.e. cross-stance attitude. These are challenges that lead to lower inter-annotator agreement and produce confusion even for humans \cite{Lai2019c,Sobhani2019a}. Moreover, relying on social media data under the terms of service of the platforms, reproducibility of some datasets is not always possible \cite{arkaitz2017Longitudinal}. There is also a need for stance detection models that also consider context, for which suitable datasets are lacking. There are also cases where concepts including sentiment, stance and emotion are conflated, with few efforts to define stance \cite{Mohammad2016b,Simaki2017a} or to experimentally prove the difference between these concepts \cite{Mohammad2017a,Aldayel2019}.

In stance particularly, we can define these problems in four levels: (1) \textbf{utterance level} as changing stance from being in favour to being against, (2) \textbf{time level} as collective stance \cite{Nguyen2012} of public pool change from highly in favour to highly against over time, (3) \textbf{domain level} where some words change its polarity from one domain to another (such as high prices indicating a favourable stance in the context of a seller but an opposing stance for customers), and (4) \textbf{cultural level} which represents stance shift between languages or various geographical locations. Indeed, use of a machine learning model training from old data may not be directly applicable to future datasets, e.g. due to suffering from domain bias, co-variate shift and concept drift. This can be cause by the nature of controversial topics and the impact of pragmatics such as time, location and ideology.

\subsection{General Challenges}

We also identified gaps in the literature that are not exclusive to stance but have significant impact in stance prediction models such as the impact of \textbf{predefined lexicon} word resolution on the model's accuracy \cite{Somasundaran2009}. This is especially true when models dependent on a lexicon fail to capture the polarity of evolving words. Research in this direction has used pre-trained word embeddings such as GloVe \cite{Pennington}, FastText \cite{Bojanowski2017}, Elmo \cite{Peters2018}, and BERT \cite{Jacob2018bert} among others which proved to mitigate the problem of polysemy though word vector representations. This is due to the fact that these models are fed news articles and web data from different sources may be inherently biased \cite{Ruder2017}. Moreover, it has been shown that variations of architecture in state-of-the-art language models can significantly impact the performance of the model in downstream tasks. Other work focuses on flipped polarity and negation \cite{Polanyi2004}. Even though embedding models consider preceding and following words of a centre word for a given sentence (context), the temporal property of the word itself and its diachronic shift from one meaning to another has not been studied in the context of stance. The identification of diachronic shift of words has however been tackled as a standalone task \cite{Fukuhara2007,Azarbonyad2017,Tahmasebi2018,Shoemark,Dubossarsky2017,Stewart2017,Hamilton2016a,Kutuzov2018,Hamilton2016,Rumshisky2017}. This is however yet to be explored in specific applications such as stance detection. This may also impact the models' performance across different domains and time frames.

The use of models developed in the field of \textbf{NLP} has been barely explored in the context of stance detection, which have been more widely studied for other tasks such as co-reference resolution \cite{Somasundaran2009} and named entity recognition \cite{Kucuk2019,Liu2013}. Previous research has however highlighted problems in this direction \cite{Lozhnikov2020,Kucuk2019,DBLP:journals/corr/abs-1811-00706,Sobhani2019a,Lai2019c,Lai2018,Simaki2017a}, which suggests that further exploration and adaptation of NLP models may be of help. Similar to most approaches for social media data, pragmatic opinions \cite{Somasundaran2009} including short opinions with few lexicon cues can negatively impact prediction performance, including hedging \cite{Somasundaran2009}, rhetorical questions \cite{Hasan2014,Mohammad2017a}, inverse polarity \cite{Mohammad2017a,skanda2017detecting}, sarcasm \cite{Hasan2014,skanda2017detecting}, all of which can have a significant impact in the classifier, especially in the case of two-way classification models.

We can summarise the challenges into two main categories:

\begin{enumerate}
\item Current deep learning models and the existence of large pre-trained embeddings can offer highly accurate results using training datasets. However, it can lead to biased results when applied to new, \textbf{unseen data}, e.g. data pertaining to a different point in time to the one seen during training. This highlights the difficulty of the task and the need to advance research in developing models that are independent of a specific use case and dataset, which can keep evolving as the data changes. Also, there is a need to develop data from different languages to mitigate the cultural biases in existing datasets. This can help detect and explain different perspectives while using specific topics to reason, compare and contrast a model's performance. This would also help further research in stance detection models that are more stable in performance. Moreover, in the case of certain languages, such as Arabic, the use of dialectical language instead of the modern standard language presents an additional challenge. More methods need to be investigated to improve a model's performance considering contextual variation (see Section \ref{ssec:contextual-variations}).

\item The existence of social media accounts run by \textbf{bots} leads to fabricated viewpoints of events. These accounts may have been created to manipulate the true view and harm specific targets (for example, businesses or people). This manipulated information can in turn have an impact on specific points in time where the bots operate, and can jeopardise the applicability of stance detection models for certain points in time (e.g. during elections where bot participation may increase) if bots are not detected and removed from the dataset.
\end{enumerate}

In summary, understanding and detecting \textbf{semantic shift} \cite{Stewart2017,Rumshisky2017,Tahmasebi2018,Shoemark} in the meaning of words has been of much interest in linguistics and related areas of research, including political science, history. However, the majority of this literature focuses their efforts on uncovering language evolution over time, with a dearth of computational research assessing its impact in context-based prediction models such as those using embedding models.  Moreover, combining a contextual knowledge using word embeddings in prediction models can help improve performance of stance detection models by leveraging their vector representations. However, current state-of-the-art research ignores the impact of contextual changes due to pragmatic factors such as social and time dimensions when building their models. This may impact a model's performance over time and can result in \textbf{outdated datasets and models}. This is due to the dependence of these models to use static data and pre-trained word embeddings to train models. While still training on data pertaining to a particular time period, models need to leverage the evolving nature of language in an unsupervised manner to keep stance detection performance stable. Temporal deterioration of models is however not exclusive to stance detection, and has been demonstrated to have an impact in other NLP tasks such as hate speech detection \cite{florio2020time}. While some social and linguistic changes may take time \cite{Hamilton} before they occur, recent literature proved that they may also occur in short periods of time \cite{Shoemark,Azarbonyad2017}. Most importantly, unlike semantic changes which capture word fluctuations over time, temporal contextual variability may occur in corpus-based predictive models.

\section{Conclusion}
\label{sec:length}

In this survey paper discuss the impact of temporal dynamics in the development of stance detection models, by reviewing relevant literature in both stance detection and temporal dynamics of social media. Our survey delves into three main factors affecting the temporal stability of stance detection models, which includes utterance, context and influence. We then discuss existing datasets and their limited capacity to enable longitudinal research into studying and capturing dynamics affecting stance detection. This leads to our discussion on research challenges needing to be tackled to further research in capturing dynamics in stance detection, which we split in two parts including core, stance-specific challenges and more general challenges.

Today's computational models are able to process big data beyond human scale, building on digital humanities and computational linguistics. This however poses a number of challenges when dealing with longitudinally-evolving data. The changes produced by societal and linguistic evolution, among others, both of which are prominent in social media platforms, have significant impact on the shift of social beliefs by means of spreading ideas. With the proliferation of historical social media data and advanced tools, we argue for the need to build models that better capture this contextual change of stance. This necessitates furthering research in the modelling of temporal dynamics of human behaviour.

Current research is largely limited to datasets covering short periods of time. Such datasets are however of special importance when one is interested in monitoring the evolution of public stance about particular topic over time. Thus, future work should consider expanding existing datasets and adapting machine learning models by focusing on maintainability and long term performance.

Our review presents an interdisciplinary viewpoint by bridging the fields of linguistics, natural language processing and digital humanity. There is a need for bridging the efforts in the field of digital humanities by relying on large historical textual corpora and introducing large scale annotated datasets. This can then benefit longitudinal analyses and contribute to advancing stance detection models that capture linguistic and societal evolution into them.

\bibliographystyle{plainnat}
\bibliography{library}

\end{document}